\documentclass[letterpaper,9pt]{article}

\usepackage[centertags]{amsmath}
\usepackage{authblk}
\usepackage{multirow}
\usepackage{amssymb}
\usepackage{graphicx}
\usepackage{listing}
\usepackage{mathrsfs}
\usepackage{amsmath,amsthm,amssymb,fontenc,color,textcomp,amsfonts,graphicx,graphics}
\usepackage[colorlinks=true]{hyperref}
\usepackage{hyperref}% http://ctan.org/pkg/hyperref
\hypersetup{colorlinks = true,linkcolor = black,citecolor = blue}
\usepackage{xcolor}
\usepackage{booktabs}
\usepackage{hyperref}
\usepackage{float}
\restylefloat{table}
\usepackage{placeins}
\usepackage{soul} %st{}

\theoremstyle{definition}

\usepackage[linesnumbered,ruled,vlined]{algorithm2e}

\usepackage{caption}
\usepackage{subcaption}

\newcommand{\be}{\begin{equation}}
\newcommand{\ee}{\end{equation}}

\DeclareMathOperator*{\argmin}{argmin}

\begin{document}   
                                    
\title{\normalsize Optimizing Sparse Generalized Singular Vectors for Feature Selection in Proximal Support Vector Machines with Application to Breast and Ovarian Cancer Detection}% for Domain Adaptation, Gene Selection, and Classification Tasks }

\author[1]{Ugochukwu O. Ugwu\thanks{Corresponding author: Ugochukwu.Ugwu@tufts.edu, uugobinnah@gmail.com}}
\author[2]{Michael Kirby}
\affil[1]{\small Department of Electrical and Computer Engineering, Tufts University, Medford, MA 02155, USA}
\affil[2]{\small Department of Mathematics, Colorado State University, CO 80523, USA}

%\author{ { Ugochukwu O. Ugwu\thanks{\,e-mail: Ugochukwu.Ugwu@colostate.edu} ~and Michael Kirby}
%\mbox{\ }\\
%{\normalsize Department of Mathematics, Colorado State University, CO  80523}}

% {Lothar Reichel\thanks{\,e-mail: reichel@math.kent.edu} \; and\; Ugochukwu O. Ugwu\thanks{\,e-mail: Ugochukwu.Ugwu@colostate.edu}}

\date{}
%\date{\normalsize{ \large Jan 2020}}

\maketitle \vspace*{-0.5cm}

\thispagestyle{empty}

%\newpage
%==================================================================
\begin{abstract}

\noindent
This paper presents approaches to compute sparse solutions of Generalized Singular Value Problem (GSVP). 
The GSVP is regularized by $\ell_1$-norm and $\ell_q$-penalty for $0<q<1$, resulting in the $\ell_1$-GSVP and $\ell_q$-GSVP formulations. The solutions of %the overdetermined and underdetermined cases of $\ell_1$-GSVP and $\ell_q$-GSVP 
these problems are determined by applying the proximal gradient descent algorithm with a fixed step size.  The inherent sparsity levels within the computed solutions are exploited for feature selection, and subsequently, binary classification with non-parallel Support Vector Machines (SVM). For our feature selection task, SVM is integrated into the $\ell_1$-GSVP and $\ell_q$-GSVP frameworks to derive the $\ell_1$-GSVPSVM and $\ell_q$-GSVPSVM variants. Machine learning applications to cancer detection are considered. We remarkably report near-to-perfect balanced accuracy across breast and ovarian cancer datasets using a few selected features.
\vspace{.3cm}

\end{abstract}
%==================================================================

\section{\large Introduction}

We are concerned with the iterative solutions of large-scale minimization problem
\be \label{GSVP}
\min_{\mathbf{z}_i\in\mathbb{R}^{m}}\frac{\|A_i\mathbf{z}_i\|^2_2}{\|A_j\mathbf{z}_i\|^2_2}, ~~i,j\in\{1,2\}, ~~i\ne j,
\ee
where $A_1\in\mathbb{R}^{\ell \times m}$, and $A_2\in\mathbb{R}^{p \times m}$. We will refer to \eqref{GSVP} as the Generalized Singular Value Problem (GSVP). Problems of this kind were first considered in \cite{CFG}, and referred to as the variational formulation of the Generalized Singular Value Decomposition \cite[GSVD]{GV}. Here, we are interested in the extrema of the objective function in \eqref{GSVP}.

In contrast to \cite{CFG}, our goal in this paper is to determine sparse solutions of \eqref{GSVP} for $\ell\ge p\gg m$ and $m\gg \ell\ge p$. In both situations, $\mathbf{z}_i$, $i=1,2$, are approximations of the generalized singular vectors corresponding to the smallest and largest generalized singular values of the GSVD of $A_i$ and $A_j$. We will exploit inherent sparsity levels in the computed solutions to carry out feature (gene) selection, and subsequently, perform binary class classification using non-parallel or proximal support vector machines \cite{MW}.

We begin with the overdetermined case, where $\ell\ge p\gg m$. Without loss of generality, let $i=1$ and $j=2$. Then the GSVD \cite{GV} of $A_1$ and $A_2$ is defined by
\begin{align}
\begin{split} \label{GSVD}
U_1^TA_1 X &= {\rm diag}[\alpha_1, \alpha_2,\dots,\alpha_m], \\ U_2^TA_2 X &= {\rm diag}[\beta_1, \beta_2,\dots,\beta_m],
\end{split}
\end{align}
where $U_1\in \mathbb{R}^{\ell \times \ell}$ and $U_2 \in \mathbb{R}^{p\times p}$ are orthogonal matrices, and their columns correspond to the left generalized singular vectors of $A_1$ and $A_2$, respectively. Throughout this paper, the superscript $^T$ denotes transposition. 

The square matrix, $X= [\mathbf{x}_1,\mathbf{x}_2,\dots,\mathbf{x}_m] \in \mathbb{R}^{m\times m}$, is nonsingular, and is shared by $A_1$ and $A_2$. The column vectors, $\mathbf{x}_k$, $k=1,2,\dots,m$, are referred to as the right generalized singular vectors of $A_1$ and $A_2$. They satisfy:
\be \label{RefGSVP} \beta_k^2 A^T_1A_1 \mathbf{x}_k = \alpha^2_kA_2^TA_2 \mathbf{x}_k, ~~k=1,2,\dots,m, \ee
where the ratios 
$\{\alpha_1/\beta_1, \alpha_2/\beta_2, \dots, \alpha_m/\beta_m \} $ are the generalized singular values of the matrix pair $\{A_1,A_2\}$ such that $\alpha_k^2+\beta_k^2=1, ~~k=1,2,\dots,m.$ The smallest generalized singular values of $\{A_i,A_j\}$ are the stationary values of \eqref{GSVP}, and the corresponding right singular vectors are the stationary vectors \cite{GV}. 

An advantage of the GSVP is that it eliminates the need to form symmetric matrices, $A^T_1A_1$ and $A^T_2A_2$ \cite{GV}. This is particularly beneficial for large-scale problems, with more columns than rows ($\ell\ge p\gg m$), as often encountered in biological settings, such as gene expression studies, where computing both matrix-matrix products is quite prohibitive, and moreso, impractical even with enormous computational power. However, when explicit formation of $A^T_1A_1$ and $A^T_2A_2$ can be afforded, and the denominator of \eqref{GSVP} is positive definite, we may refer to \eqref{GSVP} as the Generalized Eigenvalue Problem (GEP); see \cite{P}. In this case, the solutions of \eqref{GSVP} correspond to the generalized eigenvectors associated with the smallest generalized eigenvalues, $\lambda_i$ of $\{A^T_iA_i, A^T_jA_j\}$, $i,j\in\{1,2\}, i\ne j$.

The connections between the GSVP and GEP are straightforward to establish \cite{GV} for $\ell\ge p\gg m$. This follows, since the generalized eigenvalues of $\{A^T_1A_1, A^T_2A_2\}$ are the square of the generalized singular values of $\{A_1, A_2\}$, i.e., $\lambda_k = \alpha_k^2/\beta_k^2$, $k=1,2,\dots,m$. Moreover, the right generalized singular vectors of  $\{A_1, A_2\}$ are the generalized eigenvectors of $\{A^T_1A_1, A^T_2A_2\}$.

When $m\gg \ell\ge p$, the minimization problem \eqref{GSVP} is underdetermined. Therefore, establishing a connection between the GSVP and GEP becomes very challenging. This is due to the non-uniqueness of the solutions of \eqref{GSVP}. Specifically, \eqref{GSVP} can have infinitely many solutions since $N(A_i)\cap N(A_j) \neq \mathbf{0}$, where $N(M)$ is the null space of the matrix $M$ and $\mathbf{0}$ is the zero vector \cite{GRIV}. This makes determining accurate and stable optimal solutions of \eqref{GSVP} that are less sensitive to noise and errors in the data matrices infeasible. 

To remedy the challenges associated with solving \eqref{GSVP}, we introduce regularization by replacing \eqref{GSVP} with nearby problems that can be solved stably. This leads to the following regularized problems:
\be \label{RegGSVP}
\min_{\mathbf{z}_i\in\mathbb{R}^{m}}\frac{\|A_i\mathbf{z}_i\|^2_2}{\|A_j\mathbf{z}_i\|^2_2} +\delta_i\|\mathbf{z}_i\|_1, ~~i,j\in\{1,2\}, ~~i\ne j,
\ee
where $\delta_i$ is the regularization parameter that influences the accuracy and stability of the solutions of \eqref{GSVP}. It is well known that different values of $\delta_i$ have varying impact on the solutions of $\eqref{GSVP}$, and the optimal parameter choices depend on the data matrices and applications of interest. Here and below, we will refer to the first and second terms in \eqref{RegGSVP} as the {\it fidelity} and {\it regularization} terms, respectively. 

We also will refer to the minimization problems \eqref{RegGSVP} as the $\ell_1$-GSVP. The use of $\ell_1$-norm in \eqref{RegGSVP} is to promote sparsity in the computed (approximate) generalized singular vectors. As such, only the non-zero elements of $\mathbf{z}_i$ are utilized for feature selection. The sparsity enhancing property of $\ell_1$ penalty has been widely adopted in the literature; see, e.g., \cite{T,CWB}, but the novelty of our contributions lie in its application in the context of \eqref{RegGSVP}. 

It is the purpose of this paper to develop new strategies for computing sparse solutions of \eqref{GSVP}, and then discuss how sparsity in the computed solutions can be exploited to carry out feature selection and binary classification tasks. In our specific applications, such as microarray data analysis, the number of non-zero elements in the solutions of \eqref{RegGSVP} directly correspond to the most relevant and informative genes in $A_1$ and $A_2$, while genes associated with ``near-zero'' coefficients are considered irrelevant or redundant. When the solutions of \eqref{RegGSVP} are very sparse, only a small number of features will be selected. This drastically reduces the dimensionality of the data matrices, resulting in improved robustness to noise, as well as the accuracy and efficiency of subsequent analyses, such as breast and ovarian cancer classification.

The GSVD (without $\ell_1$ penalty) has been extensively employed for gene selection purposes in the literature; see, e.g., \cite{ABB,BHMA}.  In particular, a {\it gene shaving} \cite{HTEALSCBB} approach described in \cite{BHMA} focuses on selecting the most variant genes across several cancer cell lines and tumor samples. This method iteratively projects each gene onto a chosen projection direction determined by the generalized singular values. Then progressively shaves off the least variant group of genes with the smallest positive and negative projection scores. It is noteworthy that the utilization of the GSVD projections in a non-iterative manner, for gene selection and simultaneous low-rank reconstruction of gene expression datasets, was first proposed in \cite{ABB}. An extension of this idea to higher-order GSVD for gene selection is presented in, e.g., \cite{PSVA, OGA}. 

%This approach has been applied to (i) identify common (similar) to dissimilar patterns of variations across different gene expression datasets, (ii) address the problem of integrating two or more related or different large-scale biological datasets, (iii) identify, cluster, and classify functionally related genes, and (iv) identify a few statistically significant group of genes that may aid in disease pathways discovery, medical diagnosis, treatment, and drug discovery. A discussion on gene selection with higher-order GSVD is presented in, e.g., \cite{PSVA}. %\mk{The curse of dimensionality associated with gene expression data sets makes most ML models applied to them prone to overfitting - high performance on the training but poor generalization on the test data.}

We are also interested in computing approximate solutions of \eqref{GSVP} with a greater degree of sparsity than in \eqref{RegGSVP}. This is motivated by our desire to determine a parsimonious set of predictive features (genes) that will provide better insight into the biological mechanism at play in the host response to infections. To achieve this, we will regularize \eqref{GSVP} with the $\ell_q$-norm for $0<q<1$. This results in the minimization problems
\be \label{QRegGSVP}
\min_{\mathbf{z}_i\in\mathbb{R}^{m}}\frac{\|A_i\mathbf{z}_i\|^2_2}{\|A_j\mathbf{z}_i\|^2_2} +\delta_i\|\mathbf{z}_i\|_q^q, ~~i,j\in\{1,2\}, ~~i\ne j, ~~0<q<1,
\ee
where \[\|\mathbf{y}\|_q =  \left(\sum_{i = 1}^m |y_i|^q\right)^{1/q},~~ \mathbf{y} = [y_1, y_2, \dots, y_m]^T \in \mathbb{R}^m.\] Unless otherwise stated, $|y|$ denotes the absolute value of $y\in\mathbb{R}$. 

The quantity $\|\mathbf{y}\|_q$ is referred to as the $\ell_q$-norm of $\mathbf{y}$. For $0<q<1$, the mapping $\mathbf{y}\mapsto\|\mathbf{y}\|_q$ is not a norm since the triangle inequality is not satisfied. Nevertheless, this is of interest since the $\ell_q$-norm for $0<q<1$ tends to ``shrink" entries of $\mathbf{y}$ more rapidly to zero and can be more robust to outliers than the $\ell_1$-norm. %For simplicity, we will assume that for all $q>0$, the mapping $\mathbf{y}\mapsto\|\mathbf{y}\|_q$ is a norm. 
We will refer to the minimization problems \eqref{QRegGSVP} as the $\ell_q$-GSVP. The $\ell_q$-GSVP allows for $q$ to be flexibly adjusted to attain a desired sparsity level in the computed solutions. Thus, different levels of sparsity can be adapted to different datasets; see, e.g., \cite{SCSLW,SBP} for applications of $\ell_q$-norm in a closely related context. 

We note that the minimization problems \eqref{GSVP} and \eqref{RegGSVP} are {\it non-convex} optimization problems. While \eqref{GSVP} is differentiable, its regularized counterparts in \eqref{RegGSVP} and \eqref{QRegGSVP} are non-differentiable. This is due to the presence of $\ell_q$ penalty terms for $0<q\leq1$; see \cite{VSTG2,GSSTV} for related discussions when $q=1$. The main drawback for $0<q<1$ is that the minimization problems \eqref{QRegGSVP} are purely non-convex, whereas \eqref{RegGSVP} have both non-convex and convex parts. In these situations, our objective functions have differentiable and non-differentiable parts. Thus, optimization algorithms, such as Proximal Gradient Descent (PGD), can be applied to determine approximate solutions of \eqref{GSVP} and \eqref{QRegGSVP} that are sparse; see, e.g., \cite{PB, BT} for discussions.

We remark that application of PGD to \eqref{RegGSVP} and \eqref{QRegGSVP} provides the $\ell_1$-PGD-GSVP and $\ell_q$-PGD-GSVP methods, respectively. In the $\ell_q$-PGD-GSVP method, $\|\mathbf{z}_i\|_q^q$ is first approximated by a weighted $\ell_2$-norm, then the PGD algorithm applied to solve the resulting problems. Details of these methods will be provided in Sections \ref{sec:2} and \ref{sec:3}.

Our major contributions in this paper are as follows: 
\begin{enumerate}
\item We develop novel algorithms that apply the PGD to compute sparse generalized singular vectors associated with the smallest generalized singular values of the matrix pairs in \eqref{GSVP}. This leads to the $\ell_q$-PGD-GSVP method for $0<q\leq 1$. 

\item We integrate Support Vector Machines \cite[SVM]{CV} into the $\ell_q$-PGD-GSVP framework to carry out dimensionality reduction and feature selection in high-dimensional feature space, such as gene expression dataset associated with ovarian cancer. The resulting sparse technique will be referred to as the $\ell_q$-PGD-GSVPSVM method. We mention that SVM has a well-established history in the analysis of gene expression data; see, e.g., \cite{GWBV, OWSSHOSK,FCDBSH,BGLCSFAH}, and the incorporation of SVM into the framework of \eqref{GSVP} was first proposed in \cite{MW}; see Section \ref{sec:4} for details.

\item Finally, we discuss the binary classification of breast and ovarian cancer datasets described in Section \ref{sec:5}, Table \ref{Tab:1}, with proximal SVM. The classification process uses a parsimonious set of features selected by the $\ell_1$-PGD-GSVPSVM and $\ell_q$-PGD-GSVPSVM methods; see Algorithm \ref{Alg:3}. Different choices of $q$ are adapted to select a broad range of features from each dataset.  The $\ell_{0.1}$-PGD-GSVPSVM method is quite competitive and results in $100\%$ classification accuracy with ovarian cancer dataset.  This result is consistent with the most recent baseline accuracy in the literature; see, e.g., \cite{EE}. Moreover, the superiority of $\ell_{0.1}$-PGD-GSVPSVM method lies in its ability to select the least number of informative genes. Overall, our approaches are seen to handle imbalanced data effectively.
\end{enumerate}

In the following subsection, we will explore relevant literature that focuses on the application of SVM in the context of \eqref{GSVP}.

\subsection{A Review of Related Methods}

To the best of our knowledge, we are the first to develop and apply the $\ell_q$-PGD-GSVPSVM framework for the solutions of \eqref{RegGSVP} and \eqref{QRegGSVP}. However, the integration of SVM into the generalized Rayleigh quotients \eqref{GSVP} is not new. The idea was first proposed in a seminal work by Mangasarian and Wild \cite{MW} and referred to as the Generalized Eigenvalue Proximal SVM (GEPSVM). 

The GEPSVM regularizes the numerator of \eqref{GSVP} with a squared $\ell_2$-norm. This shifts the spectrum of $A_i^TA_i$ in \eqref{RefGSVP} by $\delta_i I$, where $I$ is an identity matrix of compatible size. The solutions of the GEPSVM determine two non-parallel separating hyperplanes for binary classification tasks while the GEPSVM demands that each hyperplane be closest to the samples of one class and further from the samples of the other class. The classification of new input data depends on their proximity to either of the hyperplanes. 

Several different formulations of the GEPSVM have been described in the literature; see, e.g., \cite{SCSLW, YYZYX, LSD,CY,VSTG2,GSSTV,SDCW, RLS} and references therein. In particular, the methods presented in \cite{SCSLW, YYZYX, LSD} replace the $\ell_2$-norm in \eqref{GSVP} with the $\ell_1$-norm, and solve the resulting minimization problems iteratively. These methods, commonly referred to as the Non-Parallel Proximal SVM (NPSVM), differ from the GEPSVM and the methods described in \cite{CY,VSTG2,GSSTV,SDCW, RLS} since they do not involve solving a pair of generalized eigenvalue problems. While \cite{LSD} does not employ regularization terms, \cite{SCSLW} and \cite{YYZYX} introduce $\ell_q$-norm, $q>0$, and $\ell_2$-norm regularization terms, respectively, in the numerator of the NPSVM formulation. 

Yet another closely related approach, but again different from our $\ell_1$-GSVPSVM, is the Regularized Generalized Eigenvalue Classifier \cite[ReGEC]{GCSP2} with $\ell_1$-norm regularization, proposed and applied in \cite[$\ell_1$-ReGEC]{VSTG2,GSSTV}. The ReGEC-type methods are analogous to the GEPSVM, in that, they handle singularity issues that may arise in \eqref{RefGSVP}, by shifting the spectrum of $A_i^TA_i$ and $A_j^TA_j$  by $\delta_i A_j^TA_j$ and $\delta_j A_i^TA_i$, respectively. One advantage of the ReGEC-type techniques over the GEPSVM is that only a single GEP needs to be solved to determine the two separating hyperplanes \cite{GCSP2}.

Differently from the $\ell_q$-GSVPSVM, $0<q\leq1$, both the GEPSVM and NPSVM incorporate regularization terms into the numerator of their fidelity terms.  Moreover, the practicability of the GEPSVM, NPSVM, and ReGEC approaches is significantly limited due to their enormous computational and memory requirements for large-scale problems. Specifically, their reliance on computing the generalized eigenvalue decomposition make them less attractive for large-scale problems, such as the ovarian cancer classification, where the number of columns, $m$, greatly exceeds the number of rows, $\ell$ and $p$ ($m\gg \ell\ge p$). We remark that these methods have been seen to perform exceptionally well on small-size problems for which $\ell\ge p\gg m$. Additionally, the $\ell_1$-norm sparse method proposed in \cite{PPXP} for feature selection in high-dimensional datasets requires explicit computation of $A_1^TA_1$ and $A_2^TA_2$, as well as their Cholesky factorizations.

\vspace{.3cm}
The organization of this paper is as follows. Section \ref{sec:2} provides an overview of the PGD algorithm and discusses how it can be applied to solve \eqref{RegGSVP}. This leads to the $\ell_1$-PGD-GSVP method. The $\ell_q$-PGD-GSVP method for the solutions of \eqref{QRegGSVP} is presented in Section \ref{sec:3}. This method requires approximating the $\ell_q$-norm for $0<q<1$ by a weighted $\ell_2$-norm before applying the PGD algorithm. Section \ref{sec:4} focuses on the formulation of the $\ell_q$-GSVPSVM for $0<q\leq 1$ and outlines how the PGD algorithm is employed to derive the $\ell_q$-PGD-GSVPSVM method. In Section \ref{sec:5}, numerical experiments with the breast and ovarian cancer datasets are used to illustrate the performance of the proposed methods for $\ell\ge p\gg m$ and $m\gg \ell\ge p$, respectively. Section \ref{sec:6} presents concluding remarks.

\section{\large The $\ell_1$-PGD-GSVP method for the solutions of \eqref{RegGSVP}} \label{sec:2}
This section discusses the solution method for \eqref{RegGSVP} by applying the Proximal Gradient Descent \cite[PGD]{PB} algorithm, and will be referred to as the $\ell_1$-PGD-GSVP method. We begin by briefly reviewing the PGD algorithm. 

The PGD algorithm can be applied to solve minimization problems of the form 
\[ \min_{\mathbf{z}\in\mathbb{R}^{m}} r(\mathbf{z}) + g(\mathbf{z}),\]
where $r:\mathbb{R}^m\rightarrow\mathbb{R}$ is a smooth (convex) function that is differentiable and $g:\mathbb{R}^m\rightarrow\mathbb{R}$ is a nonsmooth (convex) function that typically incorporates a regularization term. The PGD algorithm iteratively alternates between the Gradient Descent (GD) and the proximal steps. 

The GD step at the $k$th iteration is given by 
\be \label{gd}\mathbf{y}^{(k)} = \mathbf{z}^{(k)} - \alpha \nabla r(\mathbf{z}^{(k)}),\ee
where $\alpha>0$ is a step size that can be fixed or determined at each iteration using the {\it backtracking} technique; see, e.g., \cite[pg. 148]{PB} for discussion. The quantity $\nabla r(\mathbf{z}^{(k)})$ is the gradient of $r$ at $\mathbf{z}^{(k)}$. The PGD algorithm computes the gradient of $r$ at each iteration, takes a step in the direction of the steepest descent, which is given by $-\nabla r(\mathbf{z}^{(k)})$, then applies the proximal operator, denoted by $\mathbf{z}^{(k+1)}:={\bf prox}_{\alpha g}(\mathbf{y}^{(k)})$, to each iterate $\mathbf{y}^{(k)}$, in other to integrate the non-smooth function $g$ into the optimization process. The proximal operator is defined by
\be \label{proxy} {\bf prox}_{\alpha g}(\mathbf{y}^{(k)}):= \argmin_{\mathbf{z}}\left\{\|\mathbf{z}-\mathbf{y}^{(k)}\|^2_2 + g(\mathbf{z})\right\}.\ee
The goal of the proximal step \eqref{proxy} is to determine a new point $\mathbf{z}^{(k+1)}$ that is ``close" to $\mathbf{y}^{(k)}$. The PGD algorithm continues to iterate between the GD and proximal steps until a set convergence criterion is met. A notable property of the proximal operator is that $\mathbf{z}^*$ is a minimizer of $r$ if and only if $\mathbf{z}^* = {\bf prox}_{\alpha g}(\mathbf{z}^*)$.

For the minimization problems in \eqref{RegGSVP}, we let $ h(\mathbf{z}_i) =  r(\mathbf{z}_i) + g(\mathbf{z}_i)$, where \[r(\mathbf{z}_i)  = \frac{\|A_i\mathbf{z}_i\|^2_2}{\|A_j\mathbf{z}_i\|^2_2}, ~~ {\rm and}~~ g(\mathbf{z}_i) = \delta_i\|\mathbf{z}_i\|_1, ~~i,j\in\{1,2\}, ~~i\ne j.\] Since $r$ is differentiable, its gradient is given by
\be \label{ggrad}\nabla r(\mathbf{z}_i^{(k)}) = \frac{2}{\|A_j\mathbf{z}_i^{(k)}\|^2_2}\left(A_i^T(A_i\mathbf{z}_i^{(k)}) - r(\mathbf{z}_i^{(k)})A_j^T(A_j\mathbf{z}_i^{(k)})\right), ~~i,j\in\{1,2\}, ~~i\ne j.\ee Hence, the GD step \eqref{gd} is expressed as 
\be \label{ggd}\mathbf{y}^{(k)}_i = \mathbf{z}_i^{(k)} - \alpha \nabla r(\mathbf{z}_i^{(k)}), ~~i=1,2.\ee
The proximal operator that is associated with $g(\mathbf{z}_i) = \delta_i\|\mathbf{z}_i\|_1$ is readily expressed as a soft-thresholding operator. To see this, consider the explicit form of \eqref{proxy} with $g(\mathbf{z}_i) = \delta_i\|\mathbf{z}_i\|_1$. Then
\be \label{proxg}{\bf prox}_{\alpha g}(\mathbf{y}_i^{(k)})  = \argmin_{z_i}\left\{ \sum_{l=1}^m(z_{i_l}-y_{i_l}^{(k)})^2 + \alpha\delta_i |z_{i_l}| \right \},\ee
where we denote $\mathbf{x}_i\in\mathbb{R}^m$ as $\mathbf{x}_i:= [x_{i_1},x_{i_2}, \dots, x_{i_l}]^T$, $l = 1,2,\dots,m$. Let
\[f(z_{i_l}) = \sum_{l=1}^m (z_{i_l}-y_{i_l}^{(k)})^2 + \alpha\delta_i |z_{i_l}| = \sum_{l=1}^m (z_{i_l}-y_{i_l}^{(k)})^2 + \alpha\delta_i  {\rm sign}(z_{i_l})z_{i_l},\] where  ${\rm sign}(\cdot)$ is the sign operator.
Taking element-wise partial derivative of $f$ with respect to $z_{i_l}>0$ and $z_{i_l}<0$, and setting the results to zero gives
\[z_{i_l}>0: ~~~z_{i_l} = y_{i_l}^{(k)} - \frac{\alpha\delta_i}{2}, ~~~{\rm and}~~~~ z_{i_l}<0: ~~~z_{i_l} = y_{i_l}^{(k)} + \frac{\alpha\delta_i}{2}. \]
Thus, 
\be \label{pwrep}
z_{i_l} = \left\{
\begin{array}{ll}
y_{i_l}^{(k)} + \frac{\alpha\delta_i}{2}, &~~ y_{i_l}^{(k)} < -\frac{\alpha\delta_i}{2}, \\
0 &~~-\frac{\alpha\delta_i}{2}\leq y_{i_l}^{(k)} \leq \frac{\alpha\delta_i}{2},\\
y_{i_l}^{(k)} - \frac{\alpha\delta_i}{2}, &~~y_{i_l}^{(k)} > \frac{\alpha\delta_i}{2}.
\end{array}
\right.
\ee
Equation \eqref{pwrep} is a piece-wise representation of $z_{i_l}$ in terms of $y_{i_l}^{(k)}$, and it leads to the soft-thresholding operator
\be \label{thresh} \mathbf{z}_i^{(k+1)} = {\rm sign}(\mathbf{y}_i^{(k)})\cdot{\rm max}\left(|\mathbf{y}_i^{(k)}|-\frac{\alpha\delta_i}{2},0\right), ~i=1,2,\ee where $\cdot$ denotes the usual scalar multiplication.

The solution method so described is the $\ell_1$-PGD-GSVP method. Its implementation is carried out by Algorithm \ref{Alg:1}. Lines \ref{Alg:GD} and \ref{Alg:Prox} of this algorithm represent updates for the GD and proximal steps, respectively. Notice that the $\ell_1$-PGD-GSVP algorithm avoids explicit formation of $A^T_1A_1$ and $A_2^TA_2$ at each iteration to enhance computational efficiency and minimize storage requirements.

\vspace{.3cm}
\begin{algorithm}[H] 
\small %footnotesize
\SetAlgoLined
\KwIn{$A_1\in\mathbb{R}^{\ell \times m}$, $A_2\in\mathbb{R}^{\ell \times m}$, $\alpha>0$, ${\rm maxiter}$, $\delta_1>0$, $\delta_2>0$, $\mathbf{z}_1^{(0)}$, $\mathbf{z}_2^{(0)}$
}
%\KwOut{The solution $\mathbf{u}_{\epsilon,\rho}$}
\For{$k = 1:{\rm maxiter}$}{
$k:=k-1$

Compute \vspace{-.5cm} \be \label{rc} r(\mathbf{z}_i^{(k)}) = \frac{(A_i\mathbf{z}_i^{(k)})^TA_i\mathbf{z}_i^{(k)}}{(A_j\mathbf{z}_i^{(k)})^TA_j\mathbf{z}_i^{(k)}}, ~~i,j\in\{1,2\}, ~~i\ne j\ee

Compute $\nabla r(\mathbf{z}_i^{(k)})$ by using \eqref{ggrad}

Compute \label{Alg:GD} $\mathbf{y}_i^{(k)}$ by using \eqref{ggd}

Update $\label{Alg:Prox}\mathbf{z}_i^{(k+1)}$ by using \eqref{thresh}

\If{\label{cond1}$\frac{\|h(\mathbf{z}_i^{(k+1)})-h(\mathbf{z}_i^{(k)})\|_2}{\|h(\mathbf{z}_i^{(k)})\|_2}<10^{-4}$}{
break
}}
\caption{The $\ell_1$-PGD-GSVP  method for the solution of \eqref{RegGSVP}}
\label{Alg:1}
\end{algorithm}\vspace{.3cm}

\section{\large The $\ell_q$-PGD-GSVP, $0<q<1$, method for the solutions of \eqref{QRegGSVP}}\label{sec:3}

This section describes the $\ell_q$-PGD-GSVP method for the solution of non-convex minimization problems \eqref{QRegGSVP}. Our strategy is to first transform \eqref{QRegGSVP} into equivalent differentiable problems by approximating the $\ell_q$-norm with a weighted $\ell_2$-norm. Thereafter, apply the PGD algorithm to solve the resulting problems. The use of weighted $\ell_2$-norm to approximate $\ell_q$-norm for $0<q\leq 1$ is quite ubiquitous in the literature. Refer to \cite{ZXWP} for a recent discussion on the proximal operator of the $\ell_q$-norm for $0\leq q<1$. 

Let $\mathbf{z}_i = [z_{i_1}, z_{i_2}, \dots, z_{i_m}]^T$, and define $\phi_\epsilon(z_{i_k}) := \left( z_{i_k}^2 + \epsilon^2\right)^{\frac{1}{2}}$ for $\epsilon>0$. Then
\be \label{zapprox} \|\mathbf{z}_i\|^q_q \approx   \sum_{k = 1}^m \left( z_{i_k}^2 + \epsilon^2\right)^{(q-2)/2}z_{i_k}^2  = \mathbf{z}_i^T \left( \sum_{k=1}^m \phi_\epsilon(z_{i_k})^{(q-2)} \right)\mathbf{z}_i =  \mathbf{z}_i^T D_{\epsilon, q}(\mathbf{z}_i) \mathbf{z} = \|D^{\frac{1}{2}}_{\epsilon, q}(\mathbf{z}_i) \mathbf{z}_i \|_2^2,\ee
where \be \label{ddD} D_{\epsilon, q}(\mathbf{z}_i) = {\rm diag}\left(\phi_\epsilon(z_{i_1})^{(q-2)}, \phi_\epsilon(z_{i_2})^{(q-2)}, \dots, \phi_\epsilon(z_{i_m})^{(q-2)}\right) \in  \mathbb{R}^{m\times m}.\ee Substitution of \eqref{zapprox} into \eqref{QRegGSVP} yields the transformed $\ell_q$-regularized problems
\be \label{LQRegGSVP}
\min_{\mathbf{z}_i\in\mathbb{R}^{m}}\frac{\|A_i\mathbf{z}_i\|^2_2}{\|A_j\mathbf{z}_i\|^2_2} +\delta_i\|D^{\frac{1}{2}}_{\epsilon, q}(\mathbf{z}_i)\mathbf{z}_i\|^2_2, ~~i,j\in\{1,2\}, ~~i\ne j, ~~ 0<q<1.
\ee
The minimization problems \eqref{LQRegGSVP} are the equivalent versions of \eqref{QRegGSVP} that are differentiable with convex regularization parts. We will follow a similar approach described in Section \ref{sec:2} to derive the $\ell_q$-PGD-GSVP method. Here, only the derivation of the proximal step of the $\ell_q$-PGD-GSVP method will be presented while the GD step is given by \eqref{ggd}. 

For a given point $\mathbf{z}_i^{(k)}$, let $g(\mathbf{z}_i) = \delta_i\|D^{\frac{1}{2}}_{\epsilon, q}(\mathbf{z}^{(k)}_i)\mathbf{z}_i\|^2_2$.  Then the approximate solutions of  \eqref{LQRegGSVP} can be computed as 
\[\label{Lproxy}\mathbf{z}_i^{(k+1)}= \argmin_{\mathbf{z}_i\in\mathbb{R}^{m}}\left\{\|\mathbf{z}_i-\mathbf{y}_i^{(k)}\|^2_2 +\alpha\delta_i\|D^{\frac{1}{2}}_{\epsilon, q}(\mathbf{z}^{(k)}_i)\mathbf{z}_i\|^2_2\right\}, ~~i=1,2. \]
Let $s(\mathbf{z}_i) =\|\mathbf{z}_i-\mathbf{y}_i^{(k)}\|^2_2 +\alpha\delta_i\|D^{\frac{1}{2}}_{\epsilon, q}(\mathbf{z}^{(k)}_i)\mathbf{z}_i\|^2_2$. Computing the gradient of $s$ with respect to $\mathbf{z}_i$ and setting the results to zero provides the following update rule for the proximal step:
\be \label{qupdate}\mathbf{z}_i^{(k+1)} =(I+\alpha\delta_i D_{\epsilon, q}(\mathbf{z}_i^{(k)}))^{-1}\mathbf{y}^{(k)}, ~~ i=1,2. \ee
Equation \eqref{qupdate} represent the proximal steps of the $\ell_q$-PGD-GSVP method, which is described by Algorithm \ref{Alg:2} below. In our implementation of Algorithm \ref{Alg:2}, the diagonal matrices $(I+\alpha\delta_i D_{\epsilon, q}(\mathbf{z}_i^{(k)}))$ and their inverses are not formed or computed explicitly. Rather, \eqref{qupdate} are cheaply computed as element-wise products of two vectors.

\vspace{.3cm}
\begin{algorithm}[H] 
\small %footnotesize
\SetAlgoLined
\KwIn{$A_1\in\mathbb{R}^{\ell \times m}$, $A_2\in\mathbb{R}^{\ell \times m}$, $\alpha>0$, ${\rm maxiter}$, $\delta_1>0$, $\delta_2>0$, $\mathbf{z}_1^{(0)}$, $\mathbf{z}_2^{(0)}$
}
%\KwOut{The solution $\mathbf{u}_{\epsilon,\rho}$}
\For{$k = 1:{\rm maxiter}$}{
$k:=k-1$

Compute $D_{\epsilon, q}(\mathbf{z}_i^{(k)})$ by using \eqref{ddD}

Compute $r(\mathbf{z}_i^{(k)})$ by using \eqref{rc}

Compute $\nabla r(\mathbf{z}_i^{(k)})$ by using \eqref{ggrad}

Compute \label{Alg:GD} $\mathbf{y}_i^{(k)}$ by using \eqref{ggd}

Update $\label{Alg:Prox}\mathbf{z}_i^{(k+1)}$ by using \eqref{qupdate}

\If{\label{cond1}$\frac{\|h(\mathbf{z}_i^{(k+1)})-h(\mathbf{z}_i^{(k)})\|_2}{\|h(\mathbf{z}_i^{(k)})\|_2}<10^{-4}$}{
break
}}
\caption{The $\ell_q$-PGD-GSVP method for the solution of \eqref{QRegGSVP}}
\label{Alg:2}
\end{algorithm}\vspace{.3cm}

\section{\large The $\ell_q$-GSVPSVM for $0<q\leq 1$}\label{sec:4}

This section describes the integration of the SVM framework into the $\ell_q$-GSVP for $0<q\leq 1$ (cf. \eqref{RegGSVP} and \eqref{QRegGSVP}). The resulting problems will be referred to as the $\ell_q$-GSVPSVM, $0<q\leq 1$, where $q=1$ refers to both weighted and unweighted $\ell_1$ penalty.

Let $A_1 \in \mathbb{R}^{\ell\times m}$  represent a training dataset of interest with $n_1$ (Class 0) samples, and $n_2$ (Class 1) samples, in that order, such that $\ell = n_1 + n_2$. Suppose the class matrices, $C_1\in \mathbb{R}^{n_1\times m}$ and $C_2 \in \mathbb{R}^{n_2\times m}$, represent all the training data points from $A_1$ that correspond to Class 0 and Class 1, respectively.

Also, let \be \label{csvm} \tilde{C}_1 := [C_1 ~\mathbf{e}_1]\in \mathbb{R}^{n_1\times (m+1)}~~ {\rm and}~~ \tilde{C}_2 := [C_2 ~\mathbf{e}_2]\in \mathbb{R}^{n_1\times (m+1)},\ee where $\mathbf{e}_1\in \mathbb{R}^{n_1}$ and $\mathbf{e}_2\in \mathbb{R}^{n_2}$ are vectors of ones. Then the $\ell_q$-GSVPSVM for $0<q\leq 1$ can be formulated as
 \be \label{SRegGSVP} 
\min_{\mathbf{\tilde{w}}_i\in\mathbb{R}^{m}}\frac{\|\tilde{C}_i\mathbf{\tilde{w}}_i\|^2_2}{\|\tilde{C}_j\mathbf{\tilde{w}}_i\|^2_2} +\delta_i\|\mathbf{\tilde{w}}_i\|_q^q, ~~i,j\in\{1,2\}, ~~i\ne j, 0<q\leq 1.
\ee
Notably, the use of the $\ell_q$ penalty in \eqref{SRegGSVP} distinguishes our work from \cite{MW} and others described in the literature. The vectors $\mathbf{\tilde{w}}_1$ and $\mathbf{\tilde{w}}_2$ determine two non-parallel separating hyperplanes:
\be \label{hyp} \mathbf{x}^T\mathbf{w}_1 + b_1 =0, ~~~ \text{and} ~~~ \mathbf{x}^T\mathbf{w}_2 + b_2 = 0,~~\mathbf{w}_1,\mathbf{w}_2\in \mathbb{R}^{m}, \ee where $b_1,b_2 \in \mathbb{R}$ are the {\it bias} terms; and the vectors $\mathbf{w}_1$ and $\mathbf{w}_2$ represent the normal to their respective hyperplane. The first hyperplane in \eqref{hyp} is closest to the samples of Class 0 and furthest from the samples of Class 1 while the second hyperplane is closest to the samples of Class 1 and furthest from the samples of Class 0. We will apply the PGD algorithm to solve the $\ell_q$-GSVPSVM \eqref{SRegGSVP}. This leads to the $\ell_q$-PGD-GSVPSVM methods, for $0<q\leq 1$. The performance of these methods will be illustrated in Section \ref{sec:5}. Note that when Algorithm \ref{Alg:1} is employed to compute the solutions of \eqref{SRegGSVP} for $q=1$, we will refer to the resulting method as the $\ell_1$-PGD-GSVPSVM method.

Analogously to \cite{SCSLW, YYZYX, LSD,CY,MW,GCSP2,VSTG2,GSSTV,SDCW, RLS}, we will utilize $\mathbf{w}_1$ and $\mathbf{w}_2$ to define a proximal classifier that exploits the sparsity levels in both weights to carry out feature selection and binary classification. Specifically, $\mathbf{w}_1$ and $\mathbf{w}_2$ are first arranged in descending order of magnitude to obtain the vectors $\mathbf{\hat{w}}_1$ and $\mathbf{\hat{w}}_2$. The near-zero coefficients of $\mathbf{\hat{w}}_1$ and $\mathbf{\hat{w}}_2$ correspond to the less informative features, while the most significant non-zero elements indicate more relevant features. Once the most relevant features have been identified, a proximal SVM classifier is employed for binary classification. A new sample point $\mathbf{x}$ is classified to either Class 1  or Class 0  by using
\be \label{npsvm}{\rm Class}(\mathbf{x}) :=
\begin{cases}
  0 & \text{if } \frac{|\mathbf{x}^T\mathbf{w}_1 + b_1 |}{\|\mathbf{w}_1\|_2} \leq \frac{|\mathbf{x}^T\mathbf{w}_2 + b_2|}{\|\mathbf{w}_2\|_2} \\
  1 & \text{if } {\rm otherwise}.
\end{cases}
\ee Note that the non-zero coefficients of $\mathbf{\hat{w}}_1$ and $\mathbf{\hat{w}}_2$ with the highest magnitude correspond to the top features in $C_1$ and $C_2$, while features that are associated with the near-zero coefficients of $\mathbf{\hat{w}}_1$ and $\mathbf{\hat{w}}_2$ are discarded. This process of eliminating the less informative features leads to reduction in dimensions of the feature spaces of $C_1$ and $C_2$. The entire procedure so described is summarized by Algorithm \ref{Alg:3} below.

\vspace{.3cm}
\begin{algorithm}[H]
%\small
\SetAlgoLined
\KwIn{$A_1 := \begin{bmatrix} C_1\\ C_2 \end{bmatrix}$,   $A_2 := \begin{bmatrix} \hat{C}_1\\ \hat{C}_2 \end{bmatrix}$, where $C_i\in\mathbb{R}^{n_i\times m}$ and $\hat{C}_i\in\mathbb{R}^{\hat{n}_i\times m}$, $i=1,2$, are class matrices, and $A_1$ and $A_2$ are the training and validation data matrices. $n_i$, $\hat{n}_i$ denote the number of class samples, and $m$ is the number of features in each class}

Solve the minimization problems \eqref{SRegGSVP} for $\mathbf{\tilde{w}}_1$ and $\mathbf{\tilde{w}}_2$ by using Algorithms \ref{Alg:1} and \ref{Alg:2}, respectively, with $\tilde{C}_1\in\mathbb{R}^{n_1\times (m+1)}$  and $\tilde{C}_2\in\mathbb{R}^{n_2\times (m+1)}$   defined by \eqref{csvm}

Set $\mathbf{\tilde{w}}_1:= [\mathbf{w}_1^T~b_1]^T$ and $\mathbf{\tilde{w}}_2:= [\mathbf{w}_2^T~b_2]^T$

Sort \label{feat3} $\mathbf{w}_1$ and $\mathbf{w}_2$ in descending order of magnitude, and denote the sorted analogues as $\mathbf{\hat{w}}_1$ and $\mathbf{\hat{w}}_2$. Let the rank indices associated with $\mathbf{\hat{w}}_1$ and $\mathbf{\hat{w}}_2$ be denoted by $R_1$ and $R_2$, respectively

Plot $\mathbf{\hat{w}}_1$ and $\mathbf{\hat{w}}_2$ on the same graph to determine the elbow points, $(x_1,y_1)$ and $(x_2,y_2)$, respectively 

Denote the first $x_1$ and $x_2$ indices in $R_1$ and $R_2$ as $R_1(x_1)$ and $R_2(x_2)$, respectively

%Denote the top indices to be selected as $R := R_1(k) \cup R_2(k)$. The $\cup$ operator ensures that indices appearing in both $R_1(k)$ and $R_2(k)$ are not repeated

 Set entries  \label{feat6} of $\mathbf{w}_1$ and $\mathbf{w}_2$ to zero if that do not have corresponding indices in $R_1(x_1)$ and $R_2(x_2)$, respectively 
 
Use \eqref{npsvm} to classify the samples in $A_2$
\\
\caption{The $\ell_q$-PGD-GSVPSVM, $0<q\leq 1$, method for feature selection and binary classification }
\label{Alg:3}
\end{algorithm}\vspace{.3cm}

\section{\large Numerical Experiments}\label{sec:5}

In this section, all computations are carried out on a Dell computer running Windows 11 with 13th Gen Intel(R) Core(TM) i9-13900H @ 2.60GHz and 64 GB RAM. The $\ell_q$-PGD-GSVPSVM methods are implemented and analyzed using Jupyter Notebook running on VSCode. For reproducibility of the results presented herein, we set ${\tt random\_state = 42}$. The implementation of the methods can be found \url{https://github.com/Obinnah/Sparse-GSVP/blob/main/gsvp/demos/gsvp-demo.ipynb}

We will compute sparse solutions of the $\ell_q$-GSVPSVM (cf. \eqref{SRegGSVP}) by applying Algorithms \ref{Alg:1} and \ref{Alg:2}. Then utilize the computed solutions for feature selection and binary classification purposes. 

\subsection{Datasets}
 The performance of the $\ell_q$-PGD-GSVPSVM, $0<q\leq1$, methods for $\ell\gg m$ and $m\gg \ell$ will be examined using the {\it Breast Cancer} and {\it Ovarian Cancer} datasets, respectively. 
 
 The breast cancer dataset has fewer attributes than observations, and thus, corresponds to the situation where $\ell \gg m$, i.e., there are far more samples $\ell$ than features $m$. For this dataset, we will focus on accurately classifying benign and malignant breast cancer subjects. 

The ovarian cancer dataset consists of transcriptional gene expression from ovarian cancer screening. This dataset has significantly more genes $m$ than samples $\ell$, i.e., $m\gg \ell$, and it involves classifying normal and cancer subjects.  While the breast cancer dataset can be found on \url{www.kaggle.com}, the ovarian cancer dataset can be downloaded from \url{https://csse.szu.edu.cn/staff/zhuzx/Datasets.Html}. %The most recent benchmark for both datasets are shown in Table \ref{benchmark}. 
 
% \FloatBarrier
%\begin{table}[h]
%\footnotesize
%\begin{center}
%\begin{tabular}{ccccccccccccccc} 
%\cmidrule(lr){1-3}
% \multicolumn{1}{c}{Dataset}&\multicolumn{1}{c}{Accuracy $\%$}&\multicolumn{1}{c}{References}
%\\ \cmidrule(lr){1-3}
%\multirow{1}{*}{Breast Cancer}&\multirow{1}{*}{$99.12$}&\cite{FLHA}\\ 
%\multirow{1}{*}{Ovarian Cancer}&\multirow{1}{*}{$100.0$}&\cite{AI}\\ 
%[-0.01cm]
%\cmidrule(lr){1-3}
%\end{tabular} 
%\end{center} \vspace{-.5cm}
%\caption{\small Recent benchmark accuracy per dataset. }
%\label{benchmark}
%\end{table}
%\FloatBarrier\vspace{-.5cm}

%\FloatBarrier
\begin{table}[h]
\footnotesize
\begin{center}
\begin{tabular}{ccccccccccccccc} 
\cmidrule(lr){1-6}
 \multicolumn{1}{c}{Dataset}&\multicolumn{1}{c}{Size ($\ell\times m$)}&\multicolumn{1}{c}{Class}&\multicolumn{1}{c}{Training Samples}&\multicolumn{1}{c}{Validation Samples}&\multicolumn{1}{c}{Test Samples}
\\ \cmidrule(lr){1-6}
\multirow{2}{*}{Breast Cancer}&\multirow{2}{*}{$569\times30$}&\multirow{1}{*}{Benign}&\multirow{1}{*}{$249$}&\multirow{1}{*}{$64$}&\multirow{1}{*}{$44$}\\
&&\multirow{1}{*}{Malignant}&\multirow{1}{*}{$148$}&\multirow{1}{*}{$38$}&\multirow{1}{*}{$26$}\\ \cmidrule(lr){3-6}
\multirow{2}{*}{Ovarian Cancer}&\multirow{2}{*}{$253\times15154$}&\multirow{1}{*}{Normal}&\multirow{1}{*}{$63$}&\multirow{1}{*}{$19$}&\multirow{1}{*}{$9$}\\
&&\multirow{1}{*}{Cancer}&\multirow{1}{*}{$113$}&\multirow{1}{*}{$34$}&\multirow{1}{*}{$15$}\\ \cmidrule(lr){3-5}
\cmidrule(lr){1-6}
\end{tabular} 
\end{center} \vspace{-.5cm}
\caption{\small Description of the datasets for numerical experiments}
\label{Tab:1}
\end{table} 
%\FloatBarrier 
Table \ref{Tab:1} presents a detailed description of the datasets, highlighting random splits of the class samples. Specifically,  $70\%$ of the class samples are used for training, and the remaining $30\%$ are split into validation and test sets in the ratio of $60:40$ for the breast cancer dataset and $70:30$ for the ovarian cancer dataset. The validation set is used to determine the model parameters below.

\subsection{Evaluation Metrics}

The evaluation metric for the $\ell_q$-PGD-GSVPSVM methods is the Balanced Accuracy (Bal. Acc.). This metric is well-suited for datasets with imbalanced classes and is defined by
\[{\tt Balanced~ Accuracy} := \frac{1}{2}\left(\frac{TP}{TP + FN} + \frac{TN}{TN+FP} \right) = \frac{\tt Recall + Specificity}{2}, \]
\noindent
where the True Negative (TN) represents the number of $C_1$ or $\hat{C}_1$ (cf. Algorithm \ref{Alg:3}) samples that are correctly classified, and True Positive (TP) represents the number of $C_2$ or $\hat{C}_2$ (cf. Algorithm \ref{Alg:3}) samples that are correctly classified. Furthermore, False Positive (FP) corresponds to the number of incorrectly classified $C_1$ or $\hat{C}_1$ samples, while False Negative (FN) denotes the number of incorrectly classified $C_2$ or $\hat{C}_2$ samples. Note that \[ {\tt Precision} := \frac{TP}{TP+FP}\]

To measure the stability of the selected set of features across different $q$, we compute the Avg. Jaccard Similarity Index (JSI), where for any two sets $S_1$ and $S_2$, the JSI is defined as \[{\rm JSI}(S_1,S_2) :=  \frac{|S_1\cap S_2|}{|S_1\cup S_2|},\] where $|S|$ denotes the cardinality of a set $S$. Define $\mathcal{S}_q:= \{S_1,S_2,\dots,S_{N} \}$, and let the subsets $S_i$, $i=1,2,\dots,N$, denote the set of features selected for each $q$. Then the Avg. JSI is given by \[ \frac{{\rm Total ~JSI}}{{N\choose 2}} = \frac{2\sum_{1\leq i,j\leq N}{\rm JSI}(S_i,S_j)}{(N-2)!}, ~~i\neq j.\] An Avg. JSI of 1 indicates strong similarity among the subsets of features associated with each $q$. However, an Avg. JSI close to zero indicates weak similarity among subsets of features in $\mathcal{S}_q$. 
 
\subsection{Model Parameters}
Unless otherwise stated, the model parameters for the $\ell_1$-PGD-GSVPSVM, and $\ell_q$-PGD-GSVPSVM, $0<q\leq1$, methods are as follows:
\begin{enumerate}
\item The algorithms are terminated once a set convergence criterion (cf. step \ref{cond1}, Algorithm \ref{Alg:1}) is met or the maximum number of iterations ({\tt maxiter}) is reached. Here, we set ${\tt maxiter} = 10,000$ and remark that the {\tt maxiter} can affect the accuracy of the methods as well as the number of features selected.
\item The regularization parameters, $\delta_1$ and $\delta_2$, are determined using grid search over the intervals $[10^{-4},1]$ and $[2\cdot10^{-4},1]$, respectively.
\item The step size, $\alpha$, is determined using grid search over the set \[\beta\cdot\{10^{-0.5},10^{-1},10^{-1.5},10^{-2},10^{-2.5},10^{-3},10^{-3.5}\}, ~~~ \beta>0.\]
\end{enumerate}
Table \ref{regpara} shows the model parameters determined for the $\ell_q$-PGD-GSVPSVM methods. The best parameters are expected to result in sparse solutions, i.e., solutions with steep slopes characterized by non-oscillating objective function values and relative errors (cf. Figure \ref{weights} below). Generally, the methods drive the solutions to near zero, but the induced sparsity level and the observed steep slopes may depend on $q$ and other parameters, such as $\epsilon$, step size, and {\tt maxiter}. Below, we examine the sensitivity of the computed solutions to $q$ and $\epsilon$ using the ovarian cancer dataset.

%\FloatBarrier
\begin{table}[h]
\vspace{-.1cm}
\footnotesize
\begin{center}
\begin{tabular}{ccccccccccccccc} 
\cmidrule(lr){1-5}
\multicolumn{1}{c}{Method}& \multicolumn{1}{c}{Dataset}&\multicolumn{1}{c}{$\delta_1$}&\multicolumn{1}{c}{$\delta_2$}&\multicolumn{1}{c}{$\alpha$}
\\ \cmidrule(lr){1-5}
%\multirow{4}{*}{$\ell_1$-PGD-GSVPSVM}&\multirow{2}{*}{Breast Cancer}&\multirow{2}{*}{$\frac{375001}{475000}$}&\multirow{2}{*}{$\frac{187501}{237500}$}&\multirow{2}{*}{$10^{-2.5}$}\\ \\
%&\multirow{2}{*}{Ovarian Cancer}&\multirow{2}{*}{$\frac{5009}{95000}$}&\multirow{2}{*}{$\frac{2509}{47500}$}&\multirow{2}{*}{$10^{-2.5}$}\\ 
\multirow{4}{*}{$\ell_1$-PGD-GSVPSVM}&\multirow{2}{*}{Breast Cancer}&\multirow{2}{*}{$\frac{20627}{23750}$}&\multirow{2}{*}{$\frac{20627}{23750}$}&\multirow{2}{*}{$10^{-3}$}\\ \\
&\multirow{2}{*}{Ovarian Cancer}&\multirow{2}{*}{$\frac{2259}{47500}$}&\multirow{2}{*}{$\frac{2259}{47500}$}&\multirow{2}{*}{$4\cdot10^{-3}$}\\ 
\\[-0.1cm]
\cmidrule(lr){1-5}
\multirow{4}{*}{$\ell_q$-PGD-GSVPSVM, $q=1$}&\multirow{2}{*}{Breast Cancer}&\multirow{2}{*}{$3\cdot 10^{-1}$}&\multirow{2}{*}{$4\cdot 10^{-1}$}&\multirow{2}{*}{$3 \cdot 10^{-3}$}\\ \\
&\multirow{2}{*}{Ovarian Cancer}&\multirow{2}{*}{$3\cdot 10^{-3}$}&\multirow{2}{*}{$3\cdot 10^{-3}$}&\multirow{2}{*}{$10^{-0.5}$}\\ 
\\[-0.1cm]
\cmidrule(lr){1-5}
\multirow{4}{*}{$\ell_{0.1}$-PGD-GSVPSVM}&\multirow{2}{*}{Breast Cancer}&\multirow{2}{*}{$2\cdot 10^{-2}$}&\multirow{2}{*}{$2\cdot 10^{-2}$}&\multirow{2}{*}{$2.5\cdot10^{-3}$}\\ \\
&\multirow{2}{*}{Ovarian Cancer}&\multirow{2}{*}{$6\cdot 10^{-2}$}&\multirow{2}{*}{$6\cdot 10^{-2}$}&\multirow{2}{*}{$10^{-0.5}$}\\ 
\\[-0.1cm]
\cmidrule(lr){1-5}
\end{tabular} 
\end{center} \vspace{-.5cm}
\caption{\small Model parameters.}
\label{regpara}
\end{table} 

\subsection{The Sensitivity of the Solutions to $q$}
\begin{figure}[!htb] 
\centering
\minipage{.90\textwidth}
\includegraphics[width=\linewidth]{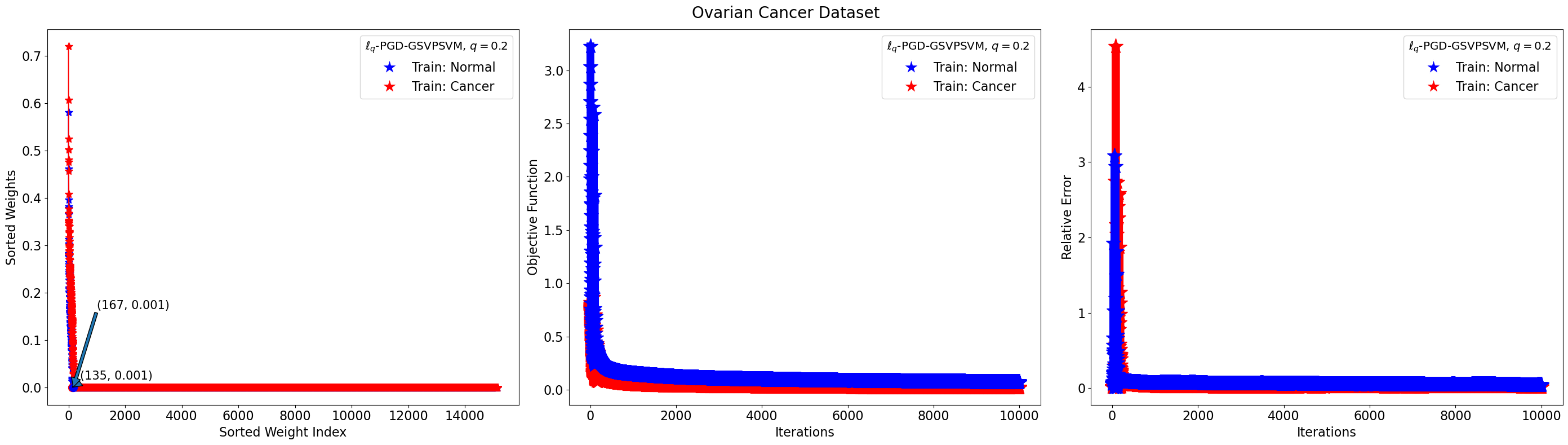} 
\endminipage%\hfill %\hspace{-1cm}

\minipage{.90\textwidth}
\includegraphics[width=\linewidth]{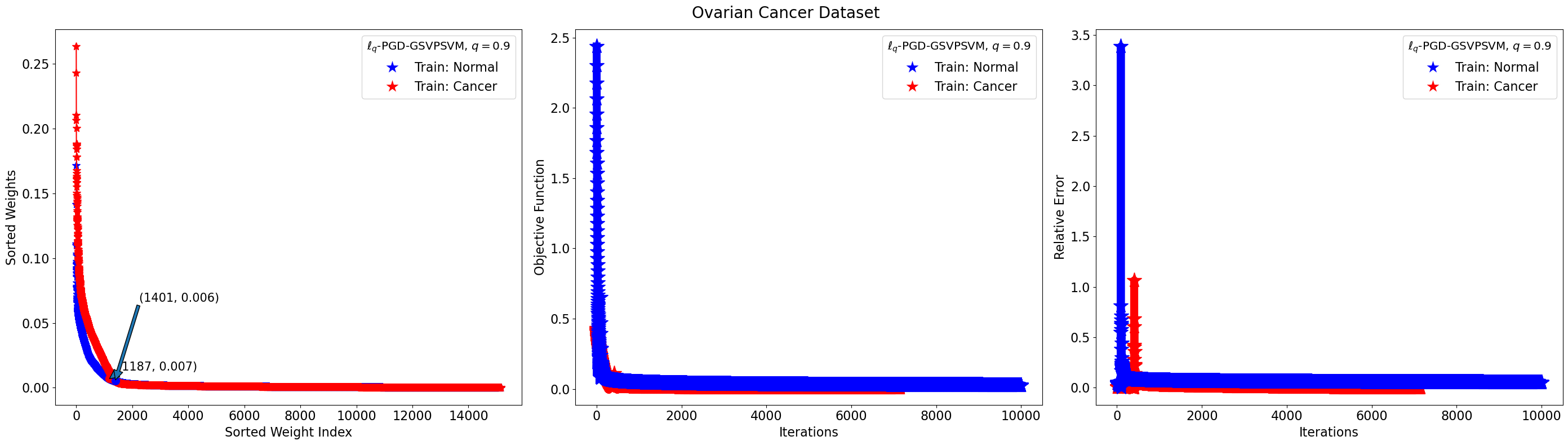} 
\endminipage%\hfill %\hspace{-1cm}
 %\hspace{-1cm} \vspace{-.05cm}
 \caption{\small The $\ell_q$-PGD-GSVPSVM method yields sparser solutions for $q=0.2$ than $q=0.9$.}
 \label{weights}
\end{figure}

\begin{figure}[!htb] 
\centering
\minipage{.90\textwidth}
\includegraphics[width=\linewidth]{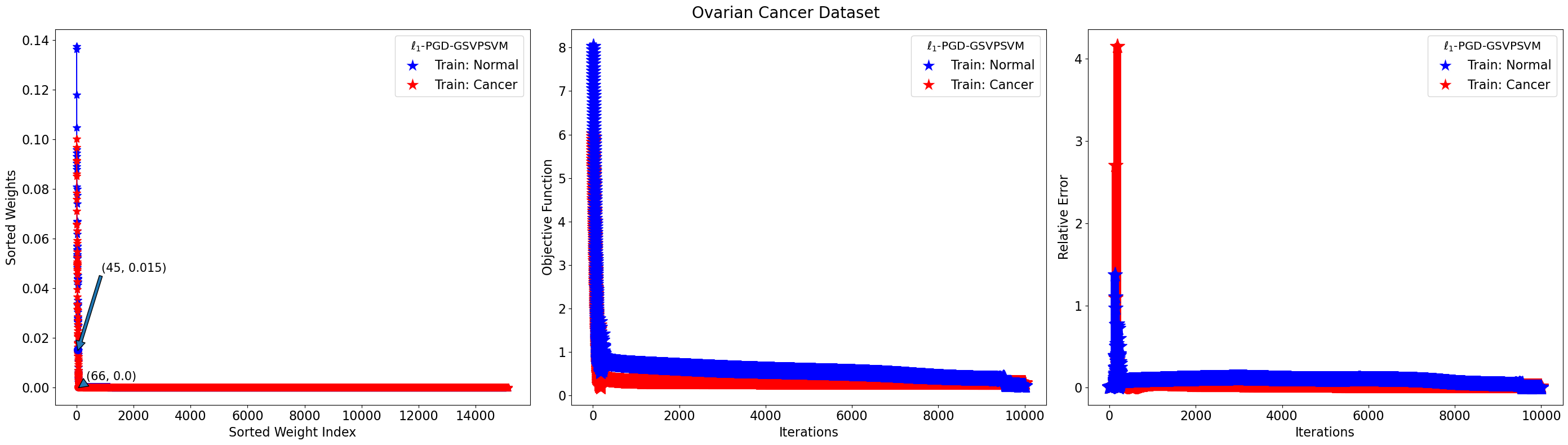} 
\endminipage%\hfill %\hspace{-1cm}
 %\hspace{-1cm} \vspace{-.05cm}
 \caption{\small The $\ell_1$-PGD-GSVPSVM method promote sparsity in the computed solutions.}
 \label{l1weights}
\end{figure}

Consider $q:=\{0.1,0.2,0.3,0.4,0.5,0.6,0.7,0.8,0.9, 1\}$ and take $\epsilon=10^{-2.5}$. We demonstrate with the ovarian cancer dataset that smaller values of $q$ often result in sparser solutions than larger values of $q$. Specifically, Figure \ref{weights} shows that the solutions determined by the $\ell_q$-PGD-GSVPSVM method for $q=0.2$ have steeper slopes than those of $q=0.9$. This behavior is consistent with other values of $q$. Analogously, Figure \ref{l1weights} shows that the $\ell_1$-PGD-GSVPSVM method also promote sparsity in the computed solutions.

\subsection{The Sensitivity of Feature Selection to $q$ and $\epsilon$}
The feature selection process is delineated in steps \ref{feat3}-\ref{feat6} of Algorithm \ref{Alg:3}. This process proceeds by first sorting the weights, $\mathbf{w}_1$ and $\mathbf{w}_2$, in descending order of magnitude, then subsequently, plots the sorted weights to identify the elbow points, as depicted in Figures \ref{weights} and \ref{l1weights}. A python library called {\tt Knee Finder}\footnote{https://pypi.org/project/kneefinder/} is employed to determine the elbow points. The determined elbow points in Figures \ref{weights} and \ref{l1weights}, represent the maximum distance from a line that connects the first and last points of each curve. The number of informative features selected for each $q$, corresponds to the smallest $x$-coordinate of the elbow points. 
 
\begin{figure}[!htb] 
\centering
\minipage{.33\textwidth}
\includegraphics[width=\linewidth]{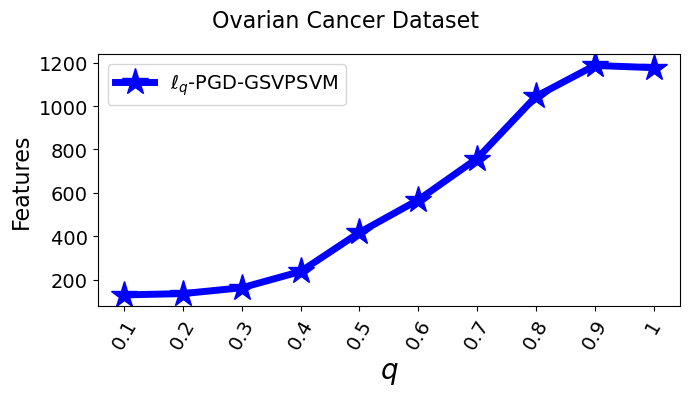} 
\endminipage
\minipage{.33\textwidth}
\includegraphics[width=\linewidth]{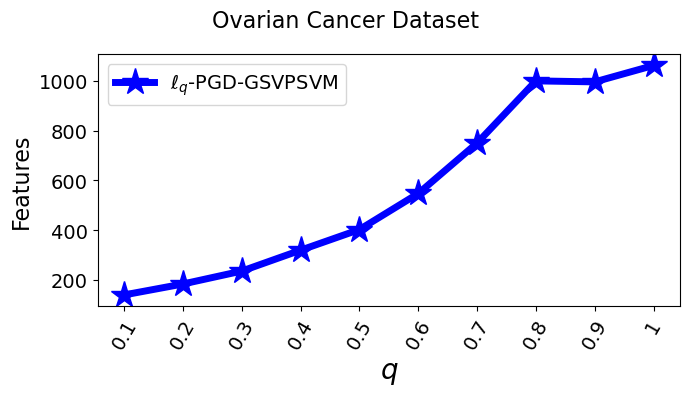} 
\endminipage
\minipage{.33\textwidth}
\includegraphics[width=\linewidth]{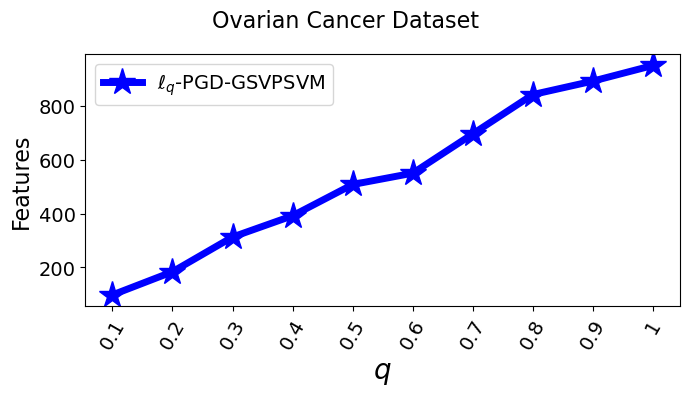} 
\endminipage

\caption{\small The number of features selected for different $q$ values with $\epsilon=10^{-2.5}$ (left), $\epsilon=10^{-2}$ (middle) and $\epsilon=10^{-1.5}$ (right).}
 \label{feats}
\end{figure} 

Figure \ref{feats} shows that the smaller the $q$, the smaller the number of features (genes) selected. This trend is consistent across different choices of $\epsilon$ considered in Figure \ref{feats}. Moreover, larger $\epsilon$ values tend to result in fewer number of features than smaller $\epsilon$ values. We report Avg. JSI of $0.6976$, $0.6303$, and $0.4260$ for the $\ell_q$-PGD-GSVPSVM, $0<q\leq 1$, method with $\epsilon=10^{-1.5}$, $\epsilon=10^{-2}$, and $\epsilon=10^{-2.5}$, respectively. These results suggest that the selected features across different $q$ exhibit greater similarity when $\epsilon = 10^{-1.5}$ is used, compared to $\epsilon=10^{-2}$ and $\epsilon=10^{-2.5}$.

%Both methods are implemented by Algorithm \ref{Alg:3}, alongside Algorithms \ref{Alg:1} and \ref{Alg:2}, respectively.  

\subsection{Model Performance on Sequestered Test Set}

We examine the performance of the $\ell_1$-PGD-GSVPSVM, and $\ell_q$-PGD-GSVPSVM, $0<q\leq1$, methods for $\ell\gg m$ and $m\gg \ell$. For simplicity, we consider $q=0.1$ and $q=1$ for weighted and unweighted $\ell_1$ penalty. The training and validation processes are carried out by Algorithm \ref{Alg:3}, alongside Algorithms \ref{Alg:1} and \ref{Alg:2}, respectively.  

\subsubsection{Case 1: $\ell \gg m$}
\begin{figure}[!htb] 
\centering
\minipage{.9\textwidth}
\includegraphics[width=\linewidth]{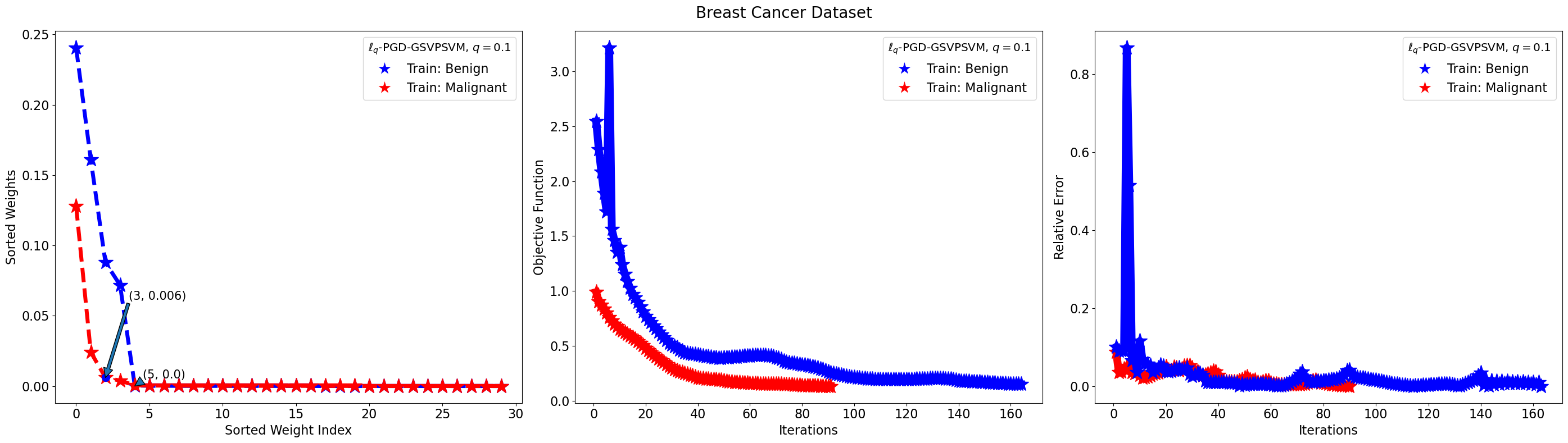} 
\endminipage
\caption{\small The solutions of the $\ell_q$-PGD-GSVPSVM method for $q=0.1$.}
 \label{sol1gsvpsvmb}
\end{figure}

We analyze the breast cancer dataset, which contains more samples $\ell$ than features $m$. Figure \ref{sol1gsvpsvmb} illustrates that the weights determined by the $\ell_{0.1}$-PGD-GSVPSVM method are sparse. Table \ref{Tab:elbowpts} presents the elbow points and selected features corresponding to the $\ell_1$-PGD-GSVPSVM and $\ell_q$-PGD-GSVPSVM ($q=0.1$ and $q=1$) methods.  These methods select $7$, $8$ and $10$ unique features, respectively. The selected features include those determined to be exclusive and common to the weights, $\mathbf{w}_1$ and $\mathbf{w}_2$. For the breast cancer dataset, the elbow points of $\mathbf{w}_1$ and $\mathbf{w}_2$ identify the most significant features for the benign and malignant cancer classes, respectively. 

\begin{figure}[!htb] 
\centering
\minipage{.90\textwidth}
\includegraphics[width=\linewidth]{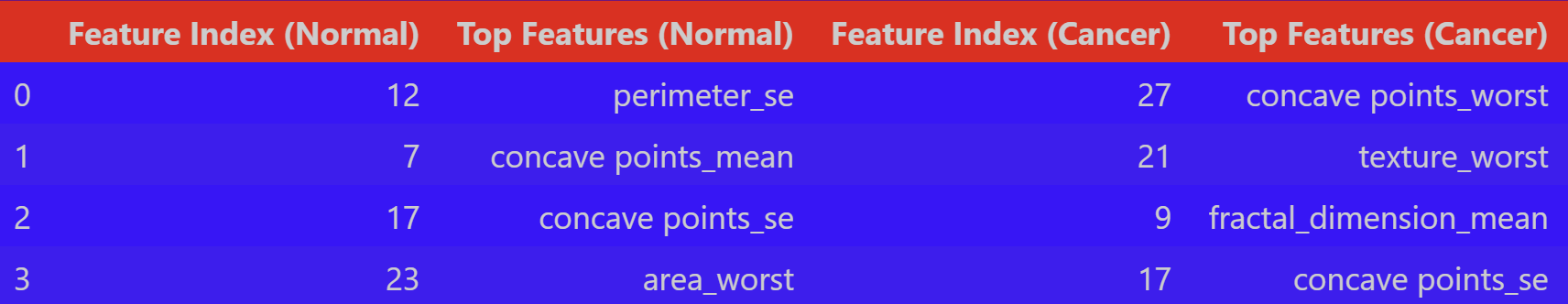} 
\endminipage

\minipage{.9\textwidth}
\includegraphics[width=\linewidth]{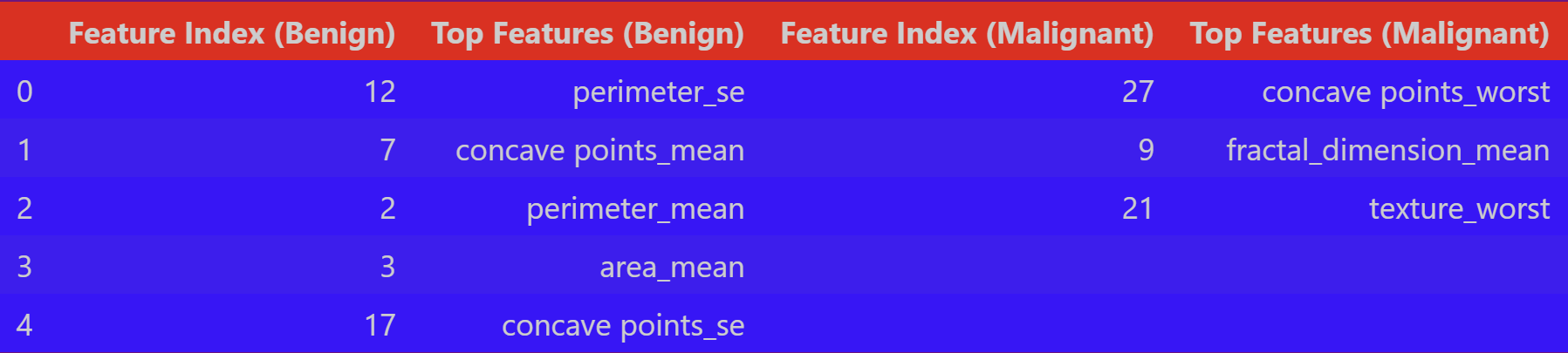} 
\endminipage

\minipage{.9\textwidth}
\includegraphics[width=\linewidth]{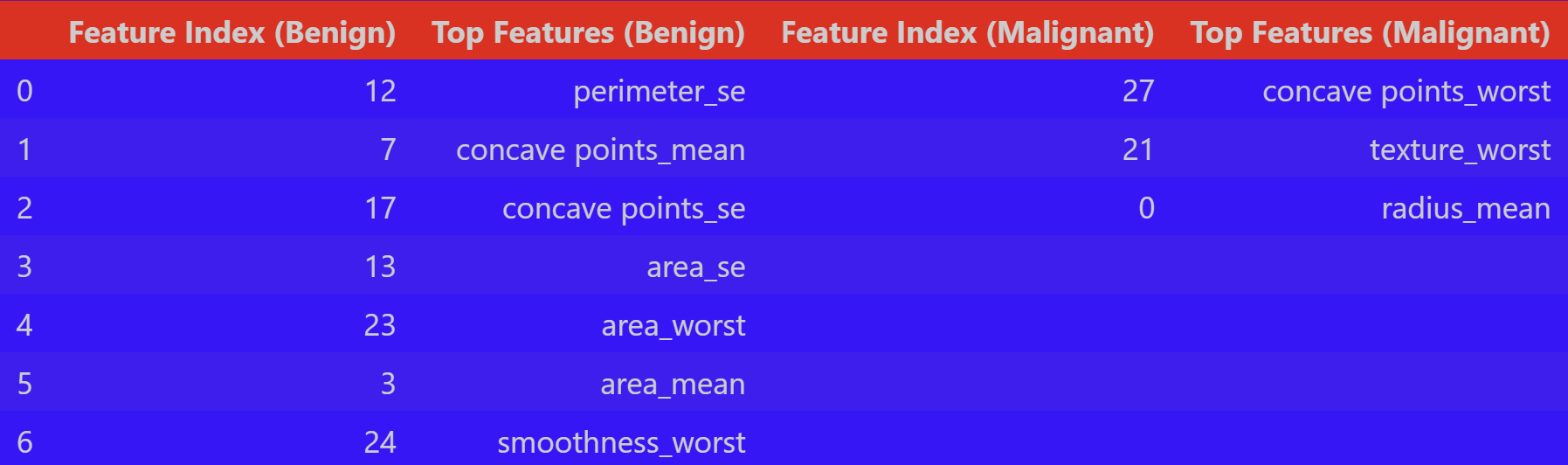} 
\endminipage
\caption{\small Top breast cancer features corresponding to  the $\ell_1$-PGD-GSVPSVM (top), and $\ell_q$-PGD-GSVPSVM methods for $q=0.1$ (middle) and $q=1$ (bottom).
}
 \label{topbreast}
\end{figure}

The most informative features corresponding to both classes are displayed in Figure \ref{topbreast} for the $\ell_1$-PGD-GSVPSVM, $\ell_{0.1}$-PGD-GSVPSVM, and $\ell_q$-PGD-GSVPSVM ($q=1$) methods. The selected features in Figure \ref{topbreast}, with the exclusion of {\tt smoothness$\_$worst} and {\tt area$\_$se}, are subsets of the 18 most significant features for breast cancer diagnosis presented in \cite{MJHMZ} using a modified recursive feature elimination approach. Using 18 features, \cite{MJHMZ} achieved a $99\%$ classification accuracy, and with only 7 features, they achieved $96\%$ accuracy. 

Table \ref{accbreast} presents the classification results for the breast cancer dataset on a sequestered test set. The $\ell_1$-PGD-GSVPSVM method yields a balanced accuracy of $96.15\%$ with only $7$ features, which is $1.13\%$ higher than those of the $\ell_q$-PGD-GSVPSVM method for $q=0.1$ and $q=1$.

\vspace{0.3cm}

\begin{table}[h]
\scriptsize
\begin{center}
\begin{tabular}{cc|c|c|ccccccccccc} 
\cmidrule(lr){1-7}
\multicolumn{1}{c}{Dataset} &\multicolumn{1}{c}{Methods} &\multicolumn{2}{|c|}{Elbow Points} & \multicolumn{3}{c}{Features Selected }  \\ 
\cmidrule(lr){3-7}
 &&$\mathbf{w}_1$ & $\mathbf{w}_2$& $\mathbf{w}_1$ (Excl.) & $\mathbf{w}_2$ (Excl.) & $\mathbf{w}_1/\mathbf{w}_2$ (Common) \\ 
\cmidrule(lr){1-7}
\multirow{3}{*}{Breast Cancer}&$\ell_1$-PGD-GSVPSVM &4 &4&3 &3 & 1 \\
&$\ell_q$-PGD-GSVPSVM $(q=1)$  &7 &3&7 & 3 & 0  \\
&$\ell_{0.1}$-PGD-GSVPSVM &5 &3&5 & 3 & 0 \\
\cmidrule(lr){1-7}
\multirow{3}{*}{Ovarian Cancer}&$\ell_1$-PGD-GSVPSVM &67 &42&60 & 35 & 7 \\
&$\ell_q$-PGD-GSVPSVM $(q=1)$ &101 &94&83 & 76 & 18  \\
&$\ell_{0.1}$-PGD-GSVPSVM &2 &2&2 & 2 & 0 \\ 
\cmidrule(lr){1-7}
\end{tabular}
\end{center} \vspace{-.5cm}
\caption{\small The elbow points and features selected by the weights of the  $\ell_1$-PGD-GSVPSVM, and $\ell_q$-PGD-GSVPSVM methods for $q=0.1$ and $q=1$.}
\label{Tab:elbowpts}
\end{table} 

%\FloatBarrier
\begin{table}[h]
\scriptsize
\begin{center}
\begin{tabular}{ccccccccccccccc} 
\cmidrule(lr){1-10}
\multicolumn{1}{c}{Dataset} &\multicolumn{1}{c}{Methods} & \multicolumn{1}{c}{Bal. Acc.} & \multicolumn{1}{c}{Specificity}& \multicolumn{1}{c}{Recall}& \multicolumn{1}{c}{Precision}& \multicolumn{1}{c}{TN}& \multicolumn{1}{c}{FP}& \multicolumn{1}{c}{FN}& \multicolumn{1}{c}{TP} \\ 
\cmidrule(lr){1-10}
\multirow{3}{*}{\shortstack{Breast Cancer \\ (Validation)}}&$\ell_1$-PGD-GSVPSVM  & 98.68 & 100& 97.37& 100& 64& 0& 1& 37 \\
&$\ell_q$-PGD-GSVPSVM $(q=1)$ & 98.68 & 100& 97.37& 100& 64& 0& 1& 37 \\
&$\ell_{0.1}$-PGD-GSVPSVM & 96.59 & 98.44& 94.74& 97.30& 63& 1& 2& 36 \\
\cmidrule(lr){1-10}
\multirow{3}{*}{\shortstack{Breast Cancer \\ (Test)}}&$\ell_1$-PGD-GSVPSVM  & 96.15 & 100& 92.31& 100& 44& 0& 2& 24 \\
&$\ell_q$-PGD-GSVPSVM $(q=1)$ & 95.02 & 97.73& 92.31& 96.00& 43& 1& 2& 24 \\
&$\ell_{0.1}$-PGD-GSVPSVM & 95.02 & 97.73& 92.31& 96.00& 43& 1& 2& 24 \\
\cmidrule(lr){1-10}
\end{tabular}
\end{center} \vspace{-.5cm}
\caption{\small  Classification reports for breast cancer dataset.}
\label{accbreast}
\end{table} 
%\FloatBarrier

\subsubsection{Case 2: $m \gg \ell$}

Here, we demonstrate the effectiveness of the $\ell_1$-PGD-GSVPSVM, $\ell_{0.1}$-PGD-GSVPSVM, and $\ell_q$-PGD-GSVPSVM ($q=1$) methods on ovarian cancer dataset, which has significantly more genes $m$ than samples $\ell$. 

\begin{figure}[!htb] 
\centering
\minipage{.98\textwidth}
\includegraphics[width=\linewidth]{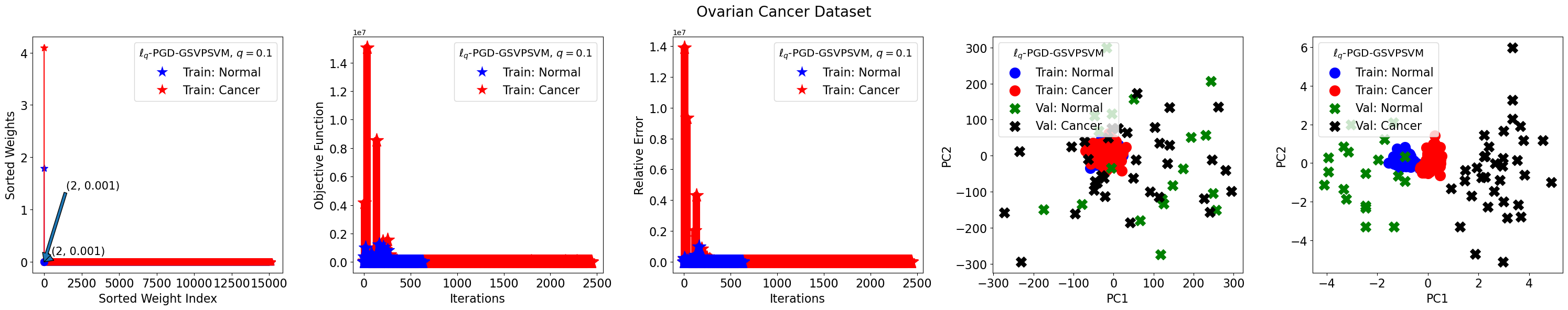} 
\endminipage

\minipage{.98\textwidth}
\includegraphics[width=\linewidth]{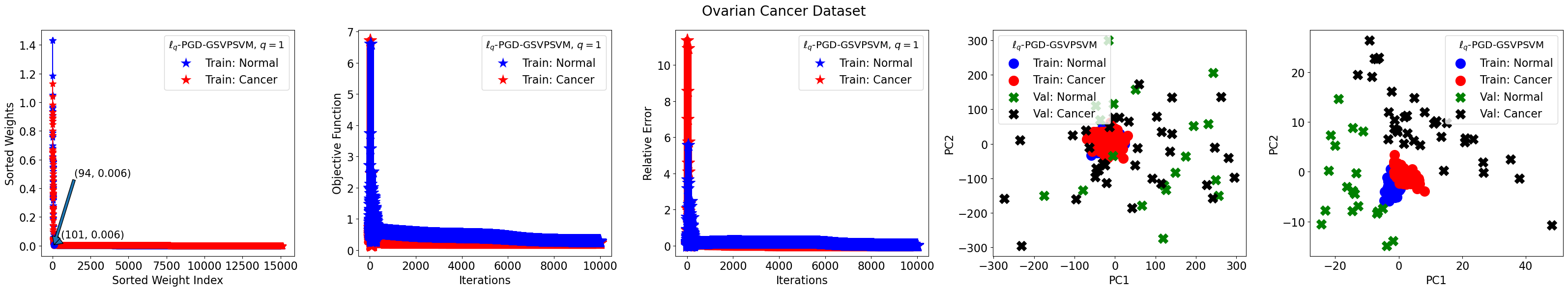} 
\endminipage

\minipage{.98\textwidth}
\includegraphics[width=\linewidth]{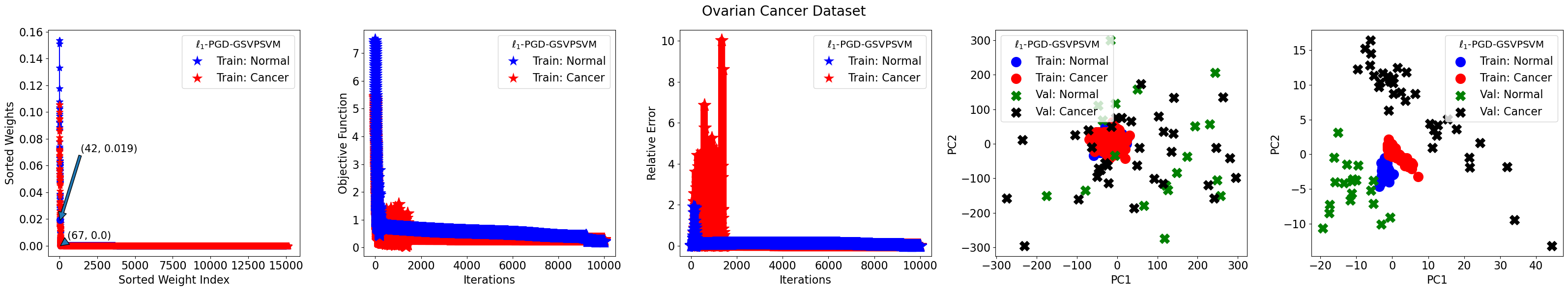} 
\endminipage

\caption{\small The solutions of the $\ell_{0.1}$-PGD-GSVPSVM, $\ell_q$-PGD-GSVPSVM ($q=1$), and $\ell_1$-PGD-GSVPSVM methods with PCA plots.}
 \label{pcafeat}
\end{figure}

\begin{figure}[!htb] 
\centering
\minipage{.9\textwidth}
\includegraphics[width=\linewidth]{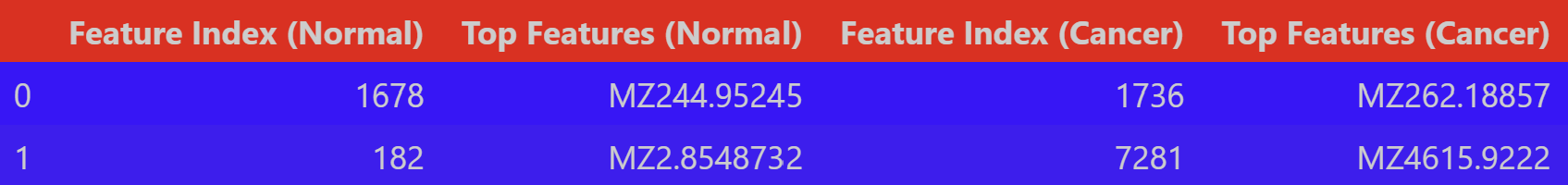} 
\endminipage

\minipage{.9\textwidth}
\includegraphics[width=\linewidth]{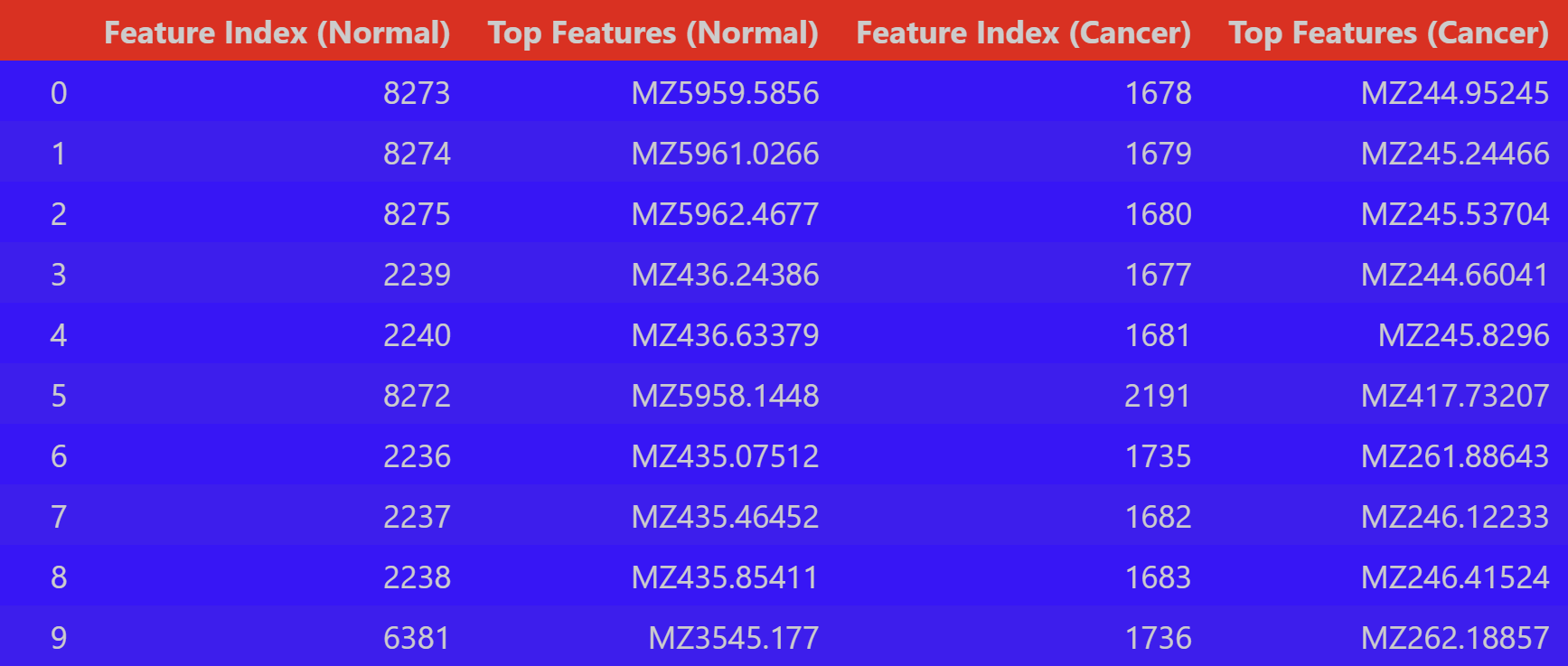} 
\endminipage

\minipage{.90\textwidth}
\includegraphics[width=\linewidth]{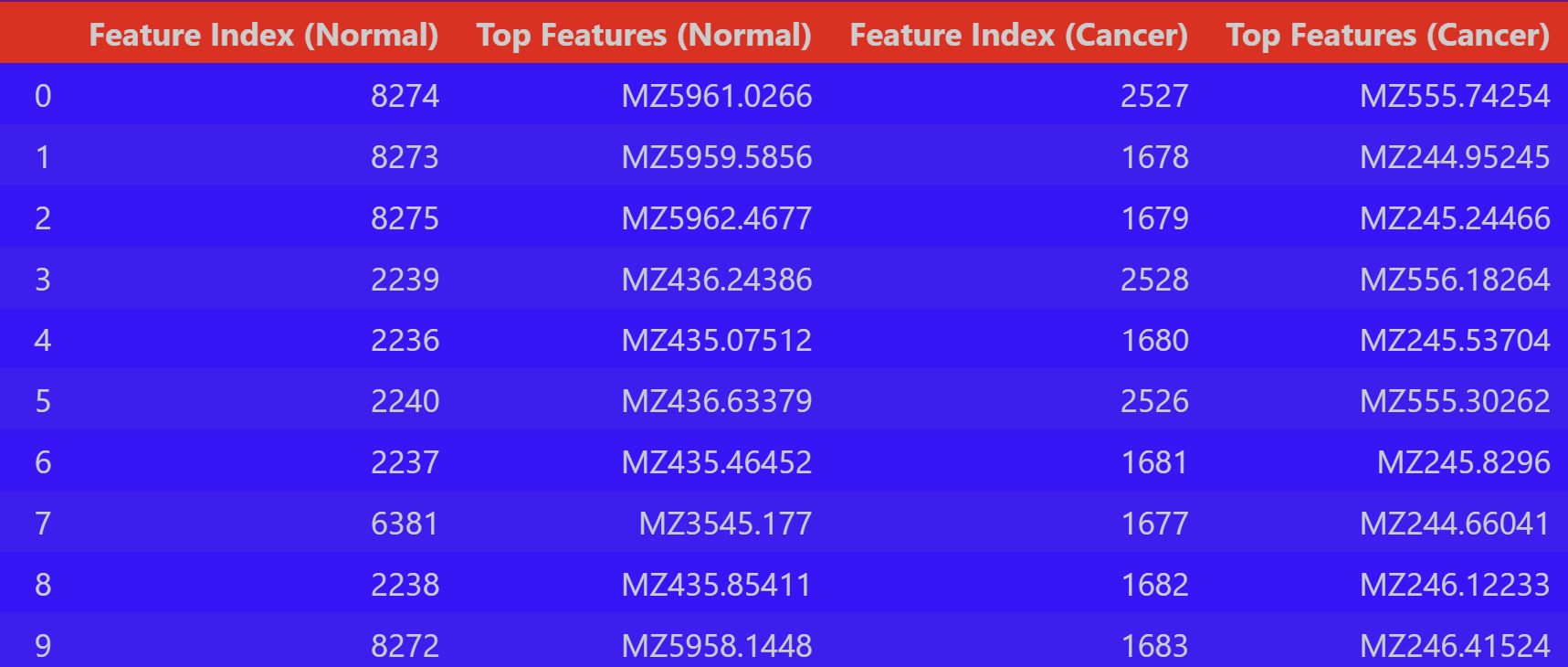} 
\endminipage

\caption{\small Top ovarian cancer genes corresponding to  the $\ell_{0.1}$-PGD-GSVPSVM (top), and $\ell_q$-PGD-GSVPSVM, $q=1$  (middle) and $\ell_1$-PGD-GSVPSVM (bottom) methods.}
 \label{feat}
\end{figure}

Figure \ref{pcafeat} illustrates that weights determined by these methods are sparse, with the  $\ell_{0.1}$-PGD-GSVPSVM method resulting in sparser weights than the other methods considered. Table \ref{Tab:elbowpts} shows that the $\ell_{0.1}$-PGD-GSVPSVM method selects 4 genes while $\ell_1$-PGD-GSVPSVM and $\ell_q$-PGD-GSVPSVM ($q=1$) methods yield $102$ and $177$ informative genes, respectively. The most significant genes selected by these methods are displayed in Figure \ref{feat}. Among the top $20$ genes selected by the $\ell_1$-PGD-GSVPSVM and $\ell_q$-PGD-GSVPSVM ($q=1$) methods, $14$ genes are consistent with the top genes selected in \cite{EE} by using filter-based methods, namely, chi-squared, F-statistic, and mutual information. 

The last two columns of Figure \ref{pcafeat} illustrate the discriminating power of the selected features. The fourth column of Figure \ref{pcafeat} shows the two-dimensional Principal Component Analysis \cite[PCA]{J} embedding of the ovarian cancer dataset using all the genes. The PC plots with the selected genes are displayed in the last column of Figure \ref{pcafeat} for the different methods. We see from Figure \ref{pcafeat}  that the selected genes can separate normal and cancer classes into distinct blobs in low-dimensional subspaces. 

\FloatBarrier
\begin{table}[h]
\scriptsize
\begin{center}
\begin{tabular}{ccccccccccccccc} 
\cmidrule(lr){1-10}
\multicolumn{1}{c}{Dataset} &\multicolumn{1}{c}{Methods} & \multicolumn{1}{c}{Bal. Acc.} & \multicolumn{1}{c}{Specificity}& \multicolumn{1}{c}{Recall}& \multicolumn{1}{c}{Precision}& \multicolumn{1}{c}{TN}& \multicolumn{1}{c}{FP}& \multicolumn{1}{c}{FN}& \multicolumn{1}{c}{TP} \\ 
\cmidrule(lr){1-10}
\multirow{3}{*}{\shortstack{Ovarian Cancer \\ (Validation)}}&$\ell_1$-PGD-GSVPSVM  & 97.06 & 100& 94.12& 100& 19& 0& 2& 32 \\
&$\ell_q$-PGD-GSVPSVM $(q=1)$ & 95.59 & 100& 91.18& 100& 19& 0& 3& 31 \\
&$\ell_{0.1}$-PGD-GSVPSVM & 100 & 100& 100& 100& 19& 0& 0& 34 \\
\cmidrule(lr){1-10}
\multirow{3}{*}{\shortstack{Ovarian Cancer \\ (Test)}}&$\ell_1$-PGD-GSVPSVM  & 96.67 & 100& 93.33& 100& 9& 0& 1& 14  \\
&$\ell_q$-PGD-GSVPSVM $(q=1)$ & 93.33 & 100& 86.67& 100& 9& 0& 2& 13 \\
&$\ell_{0.1}$-PGD-GSVPSVM & 100 & 100& 100& 100& 9& 0& 0& 15 \\
\cmidrule(lr){1-10}
\end{tabular}
\end{center} \vspace{-.5cm}
\caption{\small Classification reports for the ovarian cancer dataset.}
\label{Tab:accova}
\end{table} 
\FloatBarrier

It is noteworthy that \cite{EE} utilized $6$ genes to achieve $100\%$ classification accuracy on a test set, whereas Table \ref{Tab:accova} and Figure \ref{feat} shows that the $\ell_{0.1}$-PGD-GSVPSVM method achieved $100\%$ balanced accuracy with only $4$ genes. This demonstrates that the proposed approach can result in fewer genes while achieving superior classification accuracy.

\section{\large Conclusions}\label{sec:6}
We have demonstrated that the $\ell_q$-PGD-GSVPSVM methods for $0<q\leq1$ are viable feature selection and classification techniques that can achieve perfect to near-perfect classification accuracy. The performance of the methods are illustrated using the breast and ovarian cancer datasets. The sparsity levels in the computed solutions are exploited to select a parsimonious set of informative features. For the $\ell_q$-PGD-GSVPSVM method, smaller values of $q$ often result in fewer features being selected. More often than not, the smaller the $q$, the sparser the solutions determined by the methods. We recommend $0<q<0.4$ if a small set of features is desired.

%\vskip10pt
\vspace{2cm}
\FloatBarrier

\end{document}